\documentclass[journal]{IEEEtran} 

\usepackage{cite}
\usepackage{booktabs}
\usepackage[pdftex]{graphicx}
\usepackage{amsmath}
\usepackage{mathtools,amssymb}
\usepackage{url}
\usepackage{hyperref}
\usepackage{color}
\usepackage{multirow}
\usepackage[caption=false,font=footnotesize]{subfig}
\usepackage[normalem]{ulem}

\graphicspath{{Figures/}}

\newcommand{\note}[1]{}

\begin{document}

\title{A Comparative Study on Unsupervised Domain Adaptation 
Approaches for Coffee Crop Mapping}
%

\author{Edemir~Ferreira,
	M\'{a}rio~S.~Alvim,
	and Jefersson~A.~dos~Santos %
	\thanks{Edemir~Ferreira, M\'{a}rio~S.~Alvim, and Jefersson~A.~dos~Santos are with the Department of Computer Science, Federal University of Minas Gerais, Brazil. Emails: \{edemirm, msalvim, jefersson\}@dcc.ufmg.br}
	\thanks{The authors thank FAPEMIG (APQ-00449-17), CNPq (312167/2015-6), and CAPES for financial support.}
}

%
%

\markboth{IEEE Geoscience and Remote Sensing Letters}%
{Ferreira \MakeLowercase{\textit{et al.}}}

\maketitle

\begin{abstract}


In this work, we investigate the application of existing unsupervised domain adaptation (UDA) approaches to the task of transferring knowledge between crop regions having different coffee patterns.
Given a geographical region with fully mapped coffee plantations, we observe that this knowledge can be used to train a classifier and to map a new county with no need of samples indicated in the target region.
Experimental results show that transferring knowledge via some UDA strategies performs better than just applying a classifier trained in a region to predict coffee crops in a new one. However, UDA methods may lead to negative transfer, which may indicate that domains are too different that transferring knowledge is not appropriate.
We also verify that normalization affect significantly some UDA methods; we observe a meaningful complementary contribution between coffee crops data; and a visual behavior suggests an existent of a cluster of samples that are more likely to be drawn from a specific data.

\end{abstract}


%
\IEEEpeerreviewmaketitle

\section{Introduction}\label{intro}

Accessibility to large sets of remote sensing images (RSIs) has increased over the years, and RSIs are currently a common source of information in many agribusiness applications.
Identifying crops is essential for knowing and monitoring land use, defining new land expansion strategies, and estimating viable production value.
In this work we focus on the use of RSIs for a crucial agro-economic activity in Brazil and, in particular, the state of Minas Gerais: the coffee crop mapping.

Automatic recognition of coffee plantations using RSIs is typically modeled as a supervised classification problem.
However, this task is rather challenging, mainly because the relief and age of the crop may hinder the recognition process,
Indeed, different spectral response and texture patterns can be observed for different regions.
For instance, coffee may grow in mountainous or in flat regions (as in Brazilian cerrado-savannas), and mountains may introduce shadows and distortions in the spectral information, making the corresponding patterns appear very different from those of coffee grown in flat geographic regions. 
Because of this, spectral information may be significantly reduced or even totally lost. 
Moreover, since the growing of coffee is not a seasonal activity, there may be coffee plantations of different ages in different regions, which also affects the observed spectral patterns.

Although several approaches have advanced the state of the art in mapping coffee in recent years~\cite{SantosPenattiTorres:visapp2010,Nogueira2015:CIARP,nogueira2017towards}, one problem still remains: how to obtain representative samples for classification of new geographic areas of study?
The first possible strategy is the labeling of new samples, but it usually depends on experts, and even for them, it is not always possible to visually identify the patterns in the images. Thus, 
this process often requires visits to the study site, which can add extra costs to the analysis.
An alternative strategy to obtain extra data for training models is to transfer knowledge from already mapped regions. However, as Nogueira et al.~\cite{Nogueira2015:CIARP} has shown, due to the aforementioned differences that may exist in the coffee patterns between different geographic regions, direct transfer does not result in satisfactory quality.

In this work we investigate the application of existing unsupervised domain adaptation (UDA) approaches to the task of transferring knowledge from between crop regions having different coffee patterns. Our intent is to evaluate the effectiveness of UDA approaches to map new coffee crop areas.
UDA methods allow labeled data to be employed from one or more prior datasets with the aim of create a learning model for unseen or unlabeled data. As assumption of UDA, the source (prior labeled datasets) and target (new unlabeled data) domain have related but different probability distributions, and the divergence between such distributions is called \emph{domain shift}.
This phenomenon can arise in several visual applications, caused, for instance, by human pose-changes in estimation tasks, luminosity variations in photos, differences in acquisition sensors, or the use of multi-view descriptions of a same object (draw, sketch, photo, textual description). 

Since supervised learning methods typically expect both source and target data to follow a same distribution, the presence of domain shift can degrade the accuracy on target data if the training occurs directly in a source domain without a proper domain adaptation, i.e., a correction of the difference between source and target distributions.
Ideally, we would like to learn a proper domain adaptation in an unsupervised manner.
The task, however, is rather challenging, and its relation with realistic applications has been drawing attention in last years~\cite{zhangtransfer}.
Encouraged by these challenges, in this paper we perform a comparative experimental study of various methods for UDA in a specific view of remote sensing data. We use the dataset composed by four remote sensing images of coffee crop agriculture in scenarios with different plant and terrain conditions. 

This paper is organized as follows: Section II presents an overview of UDA techniques. Section III and IV presents, respectivelly, the evaluation protocol and experimental results of our analysis. We conclude this work in Section V with some remarks and the future directions in the research.

\section{Unsupervised Domain Adaptation Approaches}\label{related}

UDA methods can be organized in three main categories \cite{zhangtransfer}: 
(1) instance-based, (2) feature-based, and (3) classifier-based. 

In instance-based adaptation is re-weight data in the source domain or in both domains to reduce domain divergence. In feature representation is attempt to learn a new feature representations to minimize \note{decrease} domain shift and error of learning task. In classifier-based is learn a new model that minimizes the generalization error in the target domain via training instances from both domains.

In this work, we focused on feature-based UDA methods. Seven approaches were selected from the literature and are summarized in Table~\ref{properties} according to their main properties. Note that they are grouped into three branches: Data-centric, Subspace centric, and Hybrid methods. We breafly introduce each UDA method according to their branch in the next subsections.

\begin{table}[h!]
	\centering
	\caption{Summary of Unsupervised Domain Adaptation Methods}
	\label{properties}
	\tabcolsep=0.075cm
	\scalebox{0.85}
	{
		\begin{tabular}{llccccc}
			\hline
			& Approaches & \begin{tabular}[c]{@{}c@{}}Preserve\\ Variance\end{tabular} & \begin{tabular}[c]{@{}c@{}}Marginal\\ Distribution\end{tabular} & \begin{tabular}[c]{@{}c@{}}Conditional\\ Distribution\end{tabular} & \begin{tabular}[c]{@{}c@{}}Reweight\\ Instances\end{tabular} & \begin{tabular}[c]{@{}c@{}}Geometric\\ Structure\end{tabular} \\ \hline
			\multirow{3}{*}{Data Centric} & TCA~\cite{pan2011domain} & \checkmark & \checkmark &  &  &  \\
			& JDA~\cite{long2013transfer} & \checkmark & \checkmark & \checkmark &  &  \\
			& TJM~\cite{long2014transfer} & \checkmark & \checkmark &  & \checkmark &  \\ \hline
			\multirow{2}{*}{Subspace Centric} & SA~\cite{fernando2013unsupervised} & \checkmark &  &  &  &  \\
			& GFK~\cite{gong2012geodesic} & \checkmark &  &  &  & \checkmark \\ \hline
			\multirow{2}{*}{Hybrid} & CORAL~\cite{sun2016return} & \checkmark &  &  &  & \checkmark \\
			& JGSA~\cite{zhang2017joint} & \checkmark & \checkmark &  & \checkmark & \checkmark \\ \hline
		\end{tabular}
	}
\end{table}

\subsection{Data centric Approaches}

To align source and target data, data-centric methods attempt to find a specific transformation that can project both domains into a domain-invariant space. The distributional divergence between domains is reduced, while preserving the data properties from the original spaces~\cite{pan2011domain,long2013transfer,long2014transfer}. 

\noindent \textbf{Transfer Component Analysis (TCA)}~\cite{pan2011domain}: its goal is to learn a set of transfer components in a Reproducing Kernel Hilbert Space.
When projecting domain data onto the latent space spanned by the transfer components, the distance between different distributions across domains is reduced while variance is preserved.

\noindent \textbf{Joint Distribution Adaptation (JDA)}~\cite{long2013transfer}: it extends the Maximum Mean Discrepancy to measure the difference in both marginal and conditional distributions. Despite minimizing the marginal distribution between the source and target data, TCA does not guarantee that conditional distributions are reduced in this formulation, which may lead to a poor adaptation. JDA improves TCA, and integrates Maximum Mean Discrepancy with Principal Component Analysis to create a feature representation that is effective and robust for large domain shifts.

\noindent\textbf{Transfer Joint Matching (TJM)}~\cite{long2014transfer} aims at minimizing the distribution distance between domains trying to properly reweigh the instances which are more valuable to a classification task in the final adaptation. 
It is important because some instances from source data could have more relevance for classification task than others due to the huge difference of initial data representation.

TCA, JDA, and TJM rely on the assumption that there always exists a transformation function which can project the source and target data into a common subspace which, at the same time, reduce distribution difference and preserves most original information. This assumption, however, is not realistic: known problems arising from strong domain shifts suggest that there may not always exist such a space.





\subsection{Subspace Centric Approaches}

In contrast to data-centric methods, subspace-centric methods do not assume the existence of a unified transformation. They rely on a subspace manipulation of the source and target domains \cite{fernando2013unsupervised} or between them \cite{gong2012geodesic}, upholding that separate subspaces have very \note{optional} particular features to be exploited.

\noindent\textbf{Subspace Alignment (SA)}\cite{fernando2013unsupervised}: it projects source and target data onto different subspaces using PCA as a robust representation.
The method, then, learns a linear transformation matrix $M$ that aligns the source subspace to the target one while minimizing the Frobenius norm of their difference. In this way, the distance between different distributions across domains is reduced by moving closer the source and target subspaces exploiting the global covariance statistical structure of the two domains.

\noindent \textbf{Geodesic Flow Kernel (GFK)}\cite{gong2012geodesic}: is an elegant approach that integrates an infinite number of subspaces that lie on the geodesic flow from the source subspace to the target one using the kernel trick.
The main drawback of subspace-centric methods is that, while focused in reducing the geometrical shift between subspaces, the distribution shift between projected data of domains is not explicitly treated as in data-centric methods.
Moreover, the subspace dimension to project the data normally requires some tuning of parameters or preprocessing, which can be computationally expensive.


\subsection{Hybrid Approaches}

\noindent\textbf{CORAL}~\cite{sun2016return}: it was proposed to tackle the drawbacks of data and subspace-centric methods. In this approach, the domain shift is minimized by aligning the covariance of a source and target distributions in the original data.
In contrast to subspace-centric methods, CORAL suggests an alignment without the need of subspace projection, which would require intense computation and complex hyper-parameter tuning. 
In addition, CORAL do not assume a unified transformation like data-centric methods; it uses, instead, an asymmetric transformation only on source data. 

\noindent\textbf{Joint Geometric Subspace Alignment (JGSA)}~\cite{zhang2017joint}: it aims to reduce the statistical and geometrical divergence between domains using common and specific properties of the source and target data. To achieve that, an overall objective function is created by taking into account five terms: target variance, variance between/within classes, distribution shift, and subspace shift.






\section{Methodology}

We carried out an extensive set of experiments on a Brazilian Coffee Crops dataset in order to evaluate the robustness of UDA methods in a remote-sensing agriculture scenario.
The experiments were designed to answer the following research questions:

\begin{enumerate}
	\item Is UDA methods more effective than transferring without adaptation for coffee crop mapping? Which UDA approach is the most effective? When applied as a pre-processing step, how much does data normalization affect the quality of knowledge transfer?

	\item Can knowledge transfer between coffee plantations datasets from different geographic regions yield complementary results?

	\item Is it possible to infer a spatial relationship between coffee samples correctly predicted from learning models trained in different data sources?
	

\end{enumerate}

To answer question (1), we compare the selected UDA methods agaist a classifier with no adaptation strategy. We also compare the approaches by using the four most common ways to normalize data: L1-Norm, L2-Norm, L1-Norm followed by a Z-score standardize, and L2-Norm followed by a Z-score standardize. Although data pre-processing analysis is not always considered the main topic of analysis, different ways of normalizing the data before the adaptation phase can cause a great impact in the transference of knowledge.
Concerning question (2), we used Venn diagrams of predictions to analyze the complementarity among different coffee datasets according to a tuple (normalization method, UDA approach).
We selected the normalization method and UDA approach based on the most suitable tuple observed in the experiments conducted to answer question (1). 
This experiment aims at investigating the contribution of knowledge from different sources to the same target data. Since domain shift can be caused by different latent aspects in remote sensing, it is expected that different sources will have complementary contributions in consideration for the same target data.
At last, to answer (3), we perform a visual analysis of samples which are correctly predicted by specific models, using two different methods to project the original representation of data in 2D-space: Principal Component Analysis and t-Distributed Stochastic Neighbor Embedding (TSNE).

\subsection{Data}

The Brazilian Coffee Scenes~\footnote{\url{ http://www.patreo.dcc.ufmg.br/2017/11/12/brazilian-coffee-scenes-dataset/}} dataset consists of \note{comprises} four remote sensing images composed of multi-spectral scenes taken by the SPOT sensor in 2005, covering regions of coffee cultivation over four counties in the state of Minas Gerais, Brazil: Arceburgo (AR), Guaxup\'{e} (GX), Guaran\'{e}sia (GA) and Monte Santo (MS). Each county is partitioned into multiple tiles of 64 x 64 pixels, which are divided into 2 classes (coffee and non-coffee). To mitigate the problem of imbalanced datasets (which is an issue corresponding to the significant difference\note{disparity or discrepancy} among the number of samples in the different classes) we applied a random under-sampling technique, balancing the data by randomly selecting a subset of data for the targeted classes.
In our analysis, we considered each county as a different domain, thus we have four domains (AR, GX, GA and MS) leading to 12 possible domain adaptation combinations.
We have used a low-level descriptor named Border/Interior Pixel Classification (BIC) \cite{stehling2002compact} for feature extraction from coffee scenes. BIC is a very effective descriptor for coffee crops as shown in~\cite{nogueira2017towards,SantosPenattiTorres:visapp2010}.



\subsection{Setup and Implementation Details}

We made a comparison between seven state-of-the-art methods: Transfer Component Analysis (TCA) \cite{pan2011domain}, Geodesic Flow Kernel (GFK) \cite{gong2012geodesic}, Subspace Alignment (SA) \cite{fernando2013unsupervised}, Joint Distribution Analysis (JDA) \cite{long2013transfer}, Transfer Joint Matching (TJM) \cite{long2014transfer}, CORAL \cite{sun2016return}, Joint Geometrical and Statistical Alignment (JGSA) \cite{zhang2017joint} and transfer with no adaptation (NA). A brief descriptions of methods is shown in Section \ref{related} (for more details we recommend the original papers).
We follow a full training evaluation protocol, where a Support Vector Machine (SVM) is trained on the labeled source data, and tested on the unlabeled target data. In our experimental setup, tuning of parameters is always made in the source data, since is impossible to use a cross-validation without labeled samples from the target domain.
We evaluate all methods by empirically searching the parameter space for optimal parameter settings that gives the highest average kappa on all datasets, and we report the best accuracy results of each method. 

\section{Experimental Results and Discussion}

\subsection{UDA Approaches Comparison}

In this section, we compare different strategies of transferring knowledge between geographic domains in order to map coffee crops. 
We compared the SVM classifiers with no adaptation (NA) against the seven selected UDA approaches.
We also evaluated  four different way to normalize data over eight unsupervised domain adaptation approaches. In order to evaluate a normalization method, the mean accuracy value of all 12 UDA configuration was computed and reported. The mean accuracy results on each pair of counties from Brazilian Coffee Crops dataset are shown in Table \ref{tab_mean_acc}.


\begin{table}[]
	\centering
	\caption{Mean Accuracy of Methods using different Normalizations}
	\label{tab_mean_acc}
	\tabcolsep=0.155cm
	\begin{tabular}{ccccc}
		\hline
		Method & L1-Norm & L2-Norm & L1-Norm-Z-Score & L2-Norm-Z-Score \\ \hline
		NA     & 67,76   & 66,58   & 68,22           & 68,41           \\
		TCA    & 67,86   & 67,86   & \textbf{73,95}           & \textbf{74,31}           \\
		GFK    & 70,08   & 68,21   & 73,22           & 73,22           \\
		SA     & \textbf{72,80}   & 71,15   & 71,95           & 72,13           \\
		JDA    & 67,72   & 67,72   & 71,49           & 70,94           \\
		TJM    & 69,97   & 69,97   & 71,72           & 71,97           \\
		CORAL  & 64,85   & 65,38   & 71,67           & 70,94           \\
		JGSA   & 67,59   & \textbf{72,58}   & 70,18           & 70,02           \\ \hline
	\end{tabular}
\end{table}

The results show that, in general, it is better use some UDA strategy than try to transfer knowledge without no adaptation.
We can also observe that TCA achieved the best results in comparison with the other UDA approaches when using the L2 Norm followed by a Z-score standardize.
Despite the comparison of mean general results, it can be observed several remarkable points: 1) even though TCA achieving the best mean accuracy, the method had lower results in 9 of 12 combinations over the L2 Norm-Z-score setup, that could be a cost of an adaptation without taking into account the conditional distribution information; 2) the knowledge to be transferred between two domains is not always reciprocal, seems a case of JGSA which got the best results when trained at AR and test in MS but obtained the worst results when using MS to train and evaluate in AR.; 3) at the same setup mentioned before, in 3 of 12 combinations got the best results when none of the domain adaptations methods are used, that is 25\% of combinations had a negative transfer phenomenon. 
Table \ref{tab5} shows the cases of positive and negative transfer, where the UDA approaches which got a better performance in comparison with a no adaptation approach are marked as blue (Positive Transfer), otherwise they are marked with red (Negative Transfer).

\begin{table*}[]
	\centering
	\caption{Positive (blue) and Negative (red) Transfer using L2 Norm - Z-score}
	\label{tab5}
	\begin{tabular}{cccccccccccccc}
		\hline
		\begin{tabular}[c]{@{}c@{}}Method /\\ Src-Tgt\end{tabular} & AR-GX & AR-GA & AR-MS & GX-AR & GX-GA & GX-MS & GA-AR & GA-GX & GA-MS & MS-AR & MS-GX & MS-GA & Mean  \\ \hline
		NA                                                         & 66,16 & 79,17 & 63,82 & 67,03 & 66,85 & 81,31 & 69,78 & 61,60 & 65,16 & 69,78 & 85,64 & 78,44 & 68,41 \\
		TCA                                                        & \color{blue}69,75 & \color{red}75,36 & \color{blue}69,18 & \color{blue}68,13 & \color{blue}75,00 & \color{blue}82,09 & \color{blue}76,92 & \color{blue}66,30 & \color{blue}68,62 & \color{blue}73,63 & \color{red}85,08 & \color{red}77,72 & 74,31 \\
		GFK                                                        & \color{red}62,85 & \color{red}74,28 & \color{red}58,46 & \color{red}66,48 & \color{blue}73,91 & \color{blue}83,50 & \color{blue}78,02 & \color{blue}65,47 & \color{red}64,95 & \color{blue}72,53 & \color{red}85,08 & \color{red}77,54 & 73,22 \\
		SA                                                         & \color{red}61,46 & \color{red}73,91 & \color{red}56,98 & \color{red}64,84 & \color{blue}71,74 & \color{blue}83,64 & \color{blue}76,37 & \color{blue}67,54 & \color{blue}65,66 & \color{blue}72,53 & \color{red}84,53 & \color{red}76,81 & 72,13 \\
		JDA                                                        & \color{red}61,33 & \color{red}71,56 & \color{red}62,83 & \color{red}66,48 & \color{blue}77,54 & \color{red}79,27 & \color{blue}77,47 & \color{blue}66,30 & \color{blue}68,55 & \color{blue}70,33 & \color{red}84,94 & \color{red}75,18 & 70,94 \\
		TJM                                                        & \color{red}61,88 & \color{red}73,91 & \color{blue}64,74 & \color{red}64,84 & \color{blue}77,90 & \color{red}80,25 & \color{blue}74,73 & \color{blue}63,67 & \color{blue}70,03 & \color{red}69,23 & \color{red}84,81 & \color{red}77,54 & 71,97 \\
		CORAL                                                      & \color{red}63,40 & \color{red}72,10 & \color{red}61,07 & \color{red}64,29 & \color{blue}72,46 & \color{blue}82,37 & \color{blue}73,63 & \color{blue}64,23 & \color{red}63,82 & \color{red}69,78 & \color{red}85,08 & \color{red}76,81 & 70,94 \\
		JGSA                                                       & \color{red}54,01 & \color{red}65,76 & \color{blue}71,37 & \color{red}64,84 & \color{blue}81,52 & \color{red}72,92 & \color{red}51,10 & \color{blue}67,13 & \color{blue}71,93 & \color{red}68,68 & \color{red}72,24 & \color{red}72,46 & 70,02 \\ \hline
	\end{tabular}
\end{table*}

\subsection{Complementariness of Cross-Domain Predictions}

\begin{figure}[h!]
	\centering
	\subfloat[Arceburgo]{
		\includegraphics[width=0.5\columnwidth]{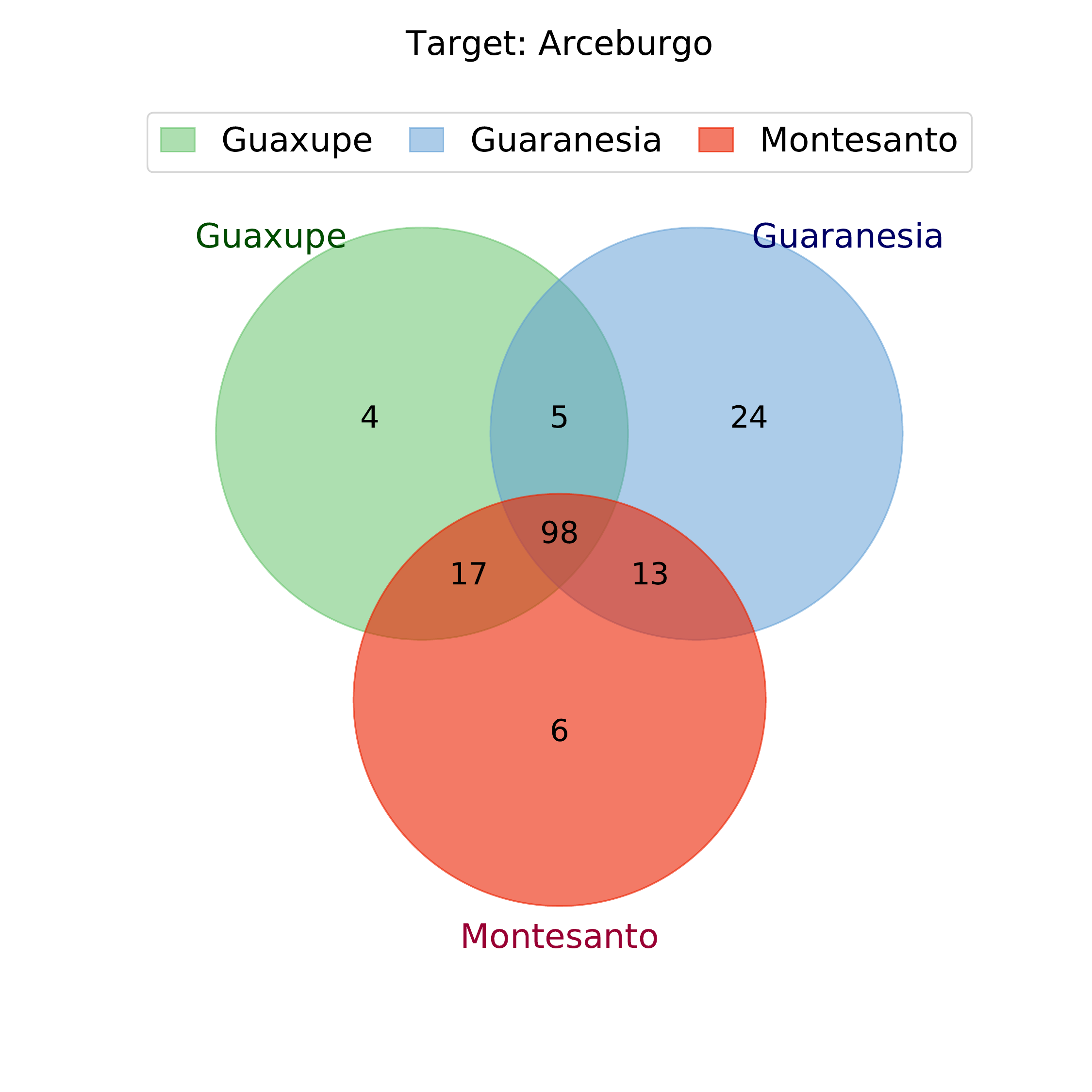}
		\label{fig:fig1}
	}
	\subfloat[Guaxup\'{e}]{
		\includegraphics[width=0.5\columnwidth]{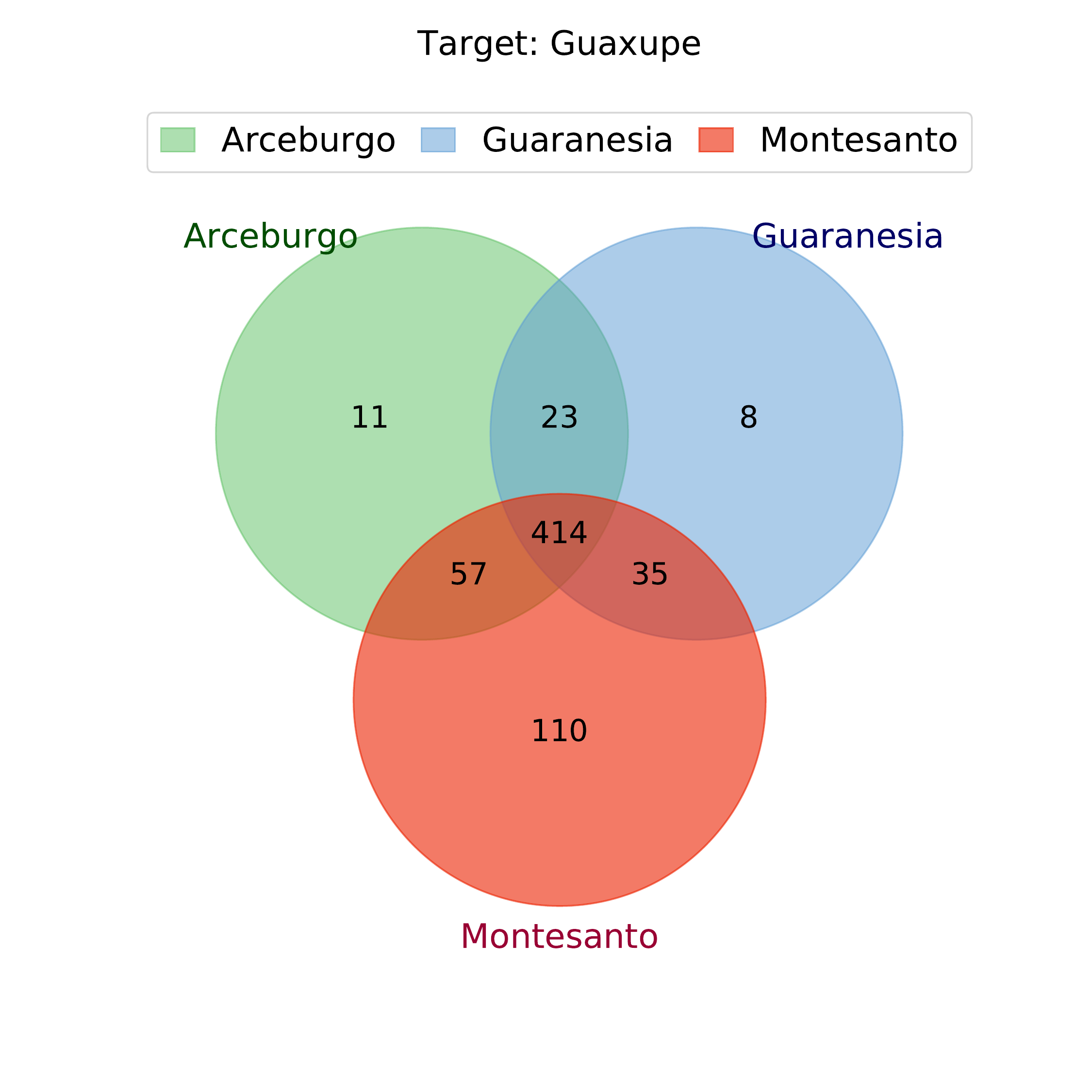}
		\label{fig:fig2}
	}
	
	\subfloat[Guaran\'{e}sia]{
		\includegraphics[width=0.5\columnwidth]{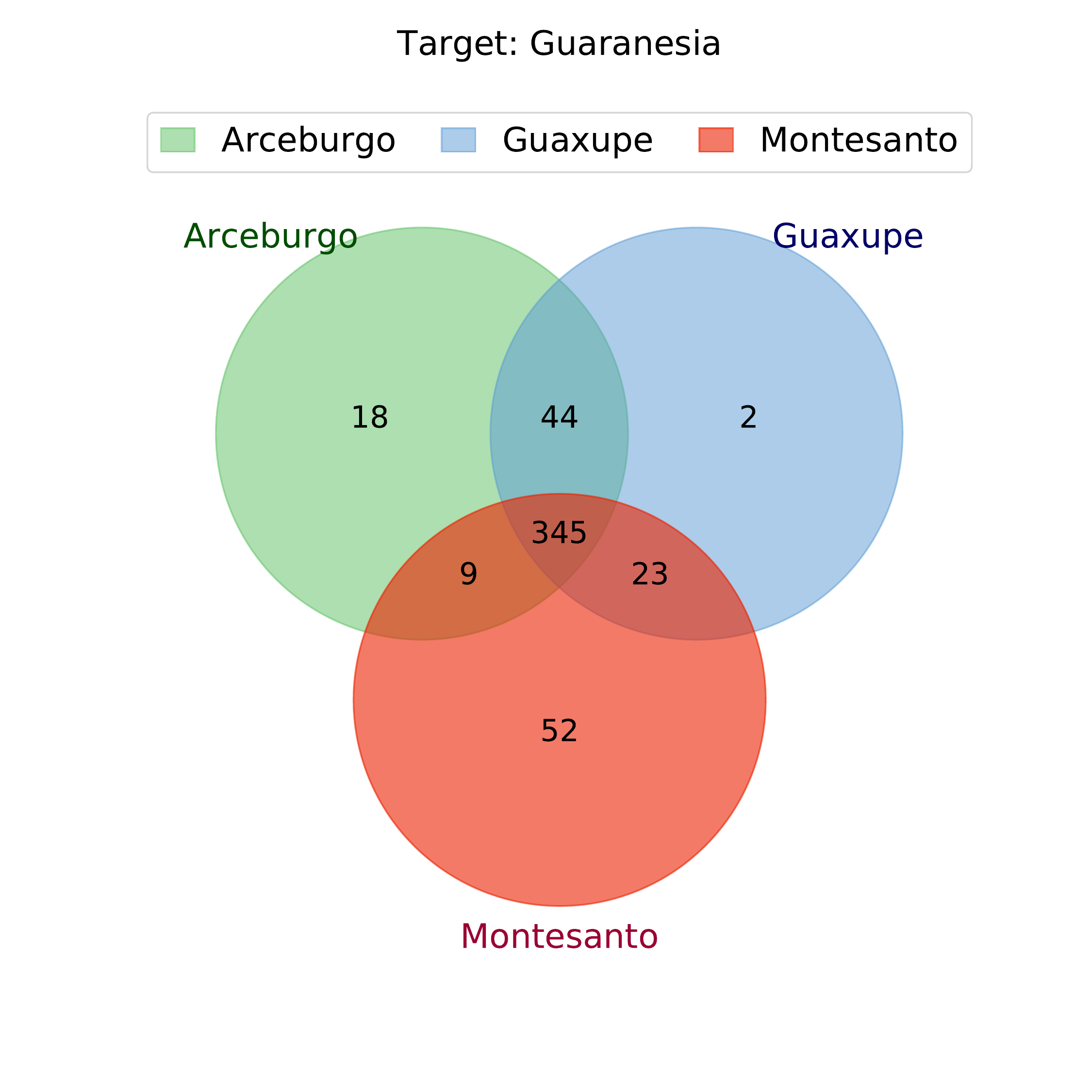}
		\label{fig:fig3}
	}
	\subfloat[Montesanto]{
		\includegraphics[width=0.5\columnwidth]{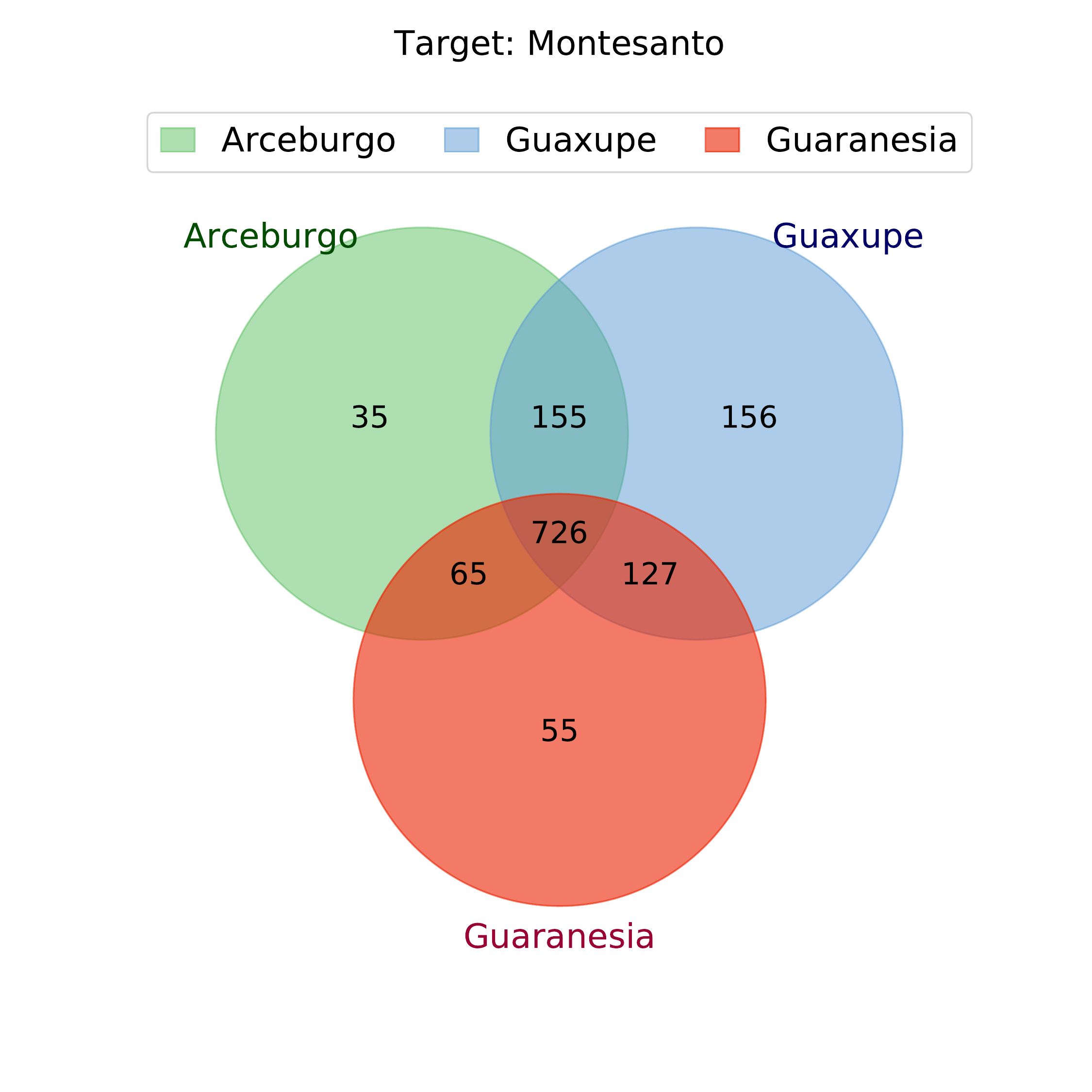}
		\label{fig:fig4}
	}
	\caption{Veen diagram for samples correctly predicted in different target data, where the intersections between sets shows samples that were predicted correctly by both sets. Target data: \textbf{(a)} Arceburgo. \textbf{(b)} Guaxup\'{e}. \textbf{(c)} Guaranesia. \textbf{(d)} Monte Santo.}
	\label{fig:venn}
\end{figure}

In this subsection, we select the pair (l2-Norm-Z-score/TCA) to analyse the complementariness of predictions between source and target data. Given a target data each group from the diagram represents a source data which the pair (l2-Norm-Z-score/TCA) was trained and afterwards test on target data. 
The intersections between sets show samples that were predicted correctly by both sets. The results were represented in Venn diagrams, which are shown in Figure \ref{fig:venn}.

As expected, most of the samples in all diagrams are at the intersection of three sets, i.e., the easiest samples are correctly predicted if trained in any of the available source datasets. However, in all cases it is possible to notice a considerable number of samples that were correctly predicted only from a single source, e.g., at Figure \ref{fig:fig2} a model trained in Montesanto is capable to correctly predict 110 samples not in common which the models trained in Arceburgo or Guaran\'{e}sia. This suggests the existence of complementary information that can be exploited to build a more reliable learning model.\note{Usar acronimos MS,AR e GA ?}
It is also noticeable a relationship of ``similarity'' between domains. That is, some pair of domains perform better than others, for instance, Guaxup\'{e} and Montesanto, in Figure \ref{fig:fig2} a model trained in Montesanto and in Figure \ref{fig:fig4} a model trained in Guaxup\'{e}. However, this relationship is not always bidirectional, for example, the case of Arceburgo and Guaran\'{e}sia, in Figure \ref{fig:fig1} Guaran\'{e}sia perform well as a source domain, predicting correctly 140 over 182 (76,92\%) samples, but in Figure \ref{fig:fig3} Montesanto is more useful than Arceburgo, predicting correctly 429 over 552 (77,71\%) samples in a comparison of 416 (75,36\%) from Arceburgo.

\subsection{Visual Analysis}

\begin{figure}[]
	\centering
	\subfloat[Arceburgo]{
		\includegraphics[width=0.48\columnwidth]{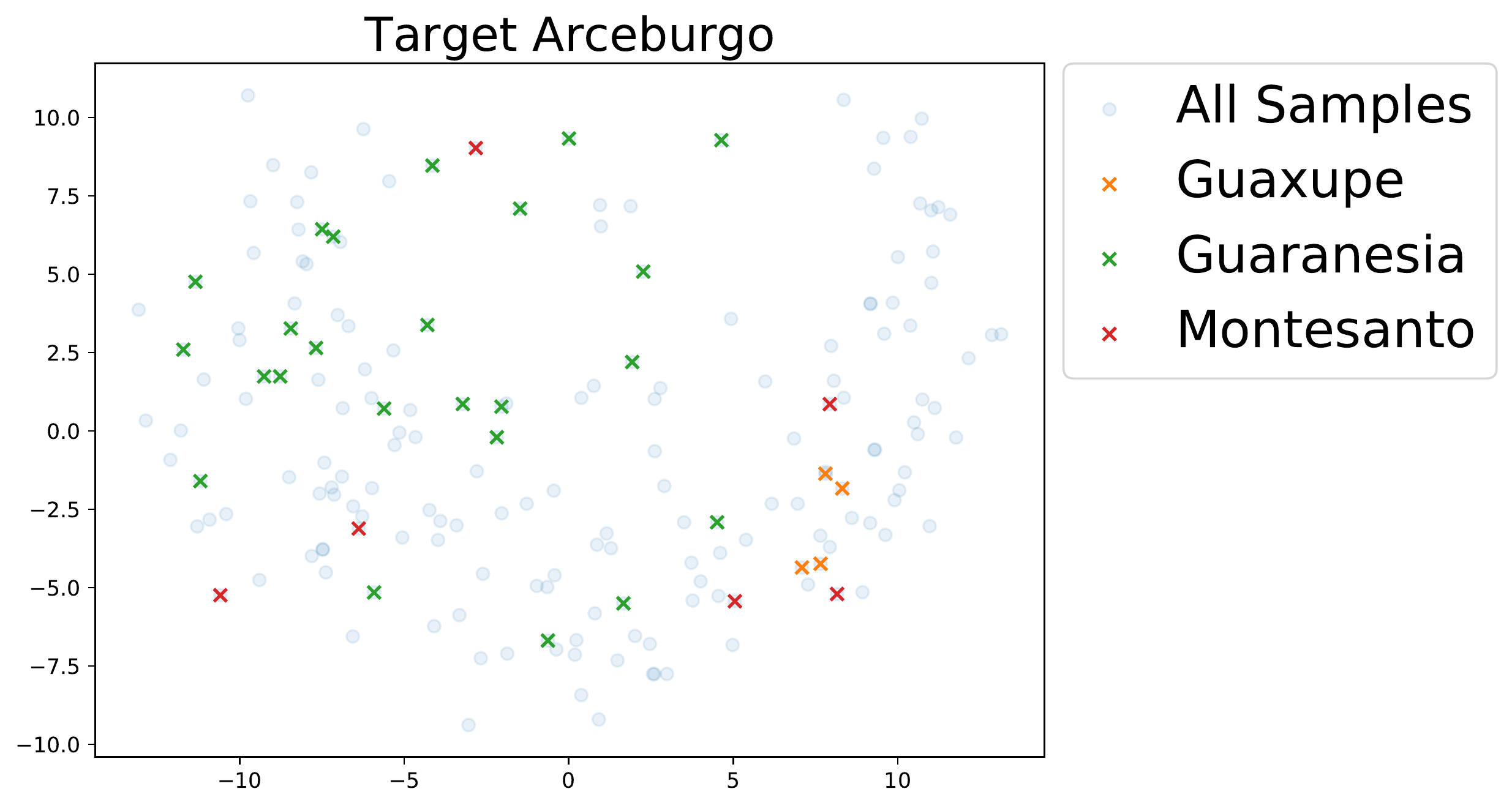}
	}
	\subfloat[Guaxup\'{e}]{
		\includegraphics[width=0.48\columnwidth]{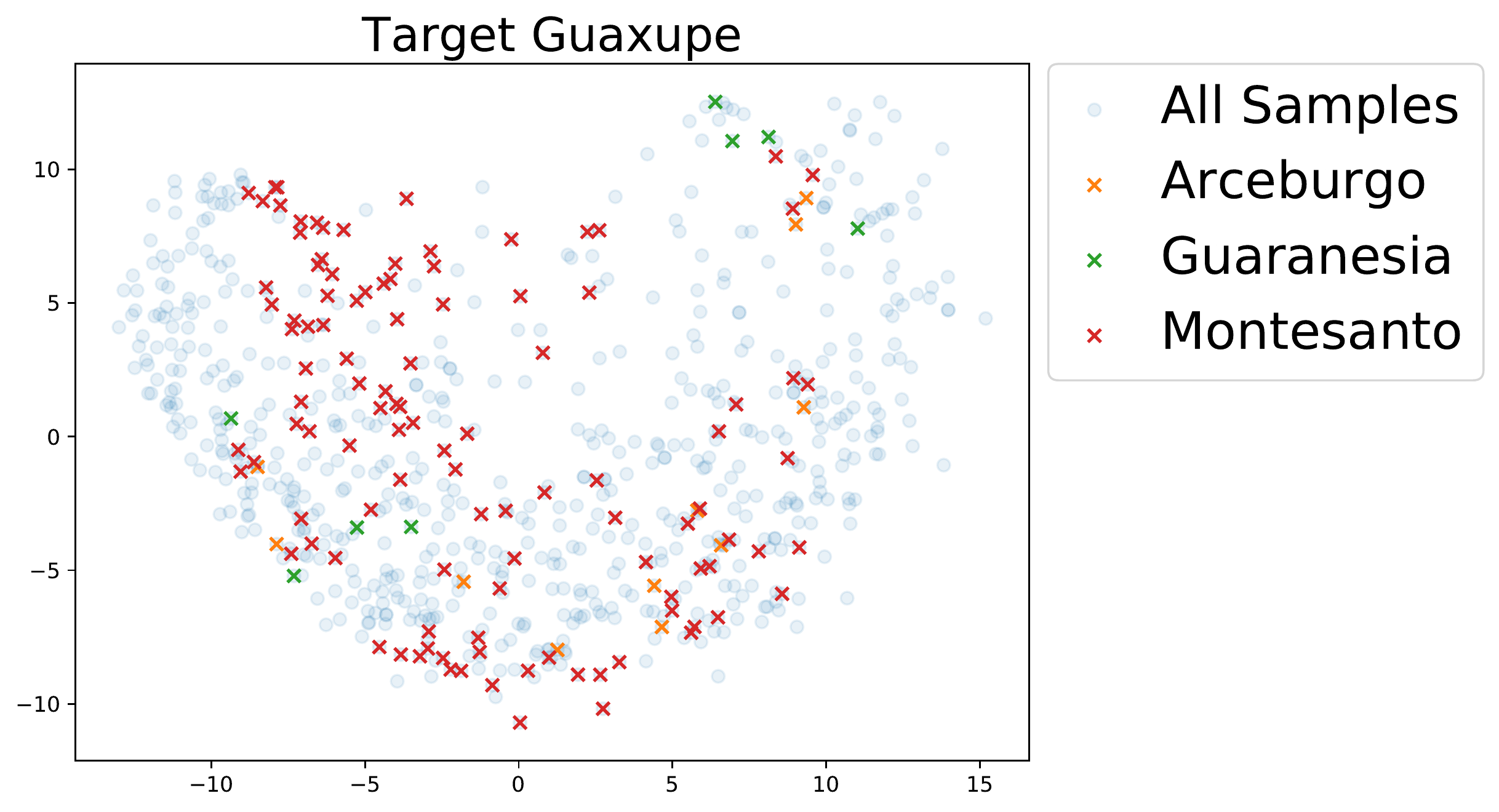}
	}

	\subfloat[Guaran\'{e}sia]{
		\includegraphics[width=0.48\columnwidth]{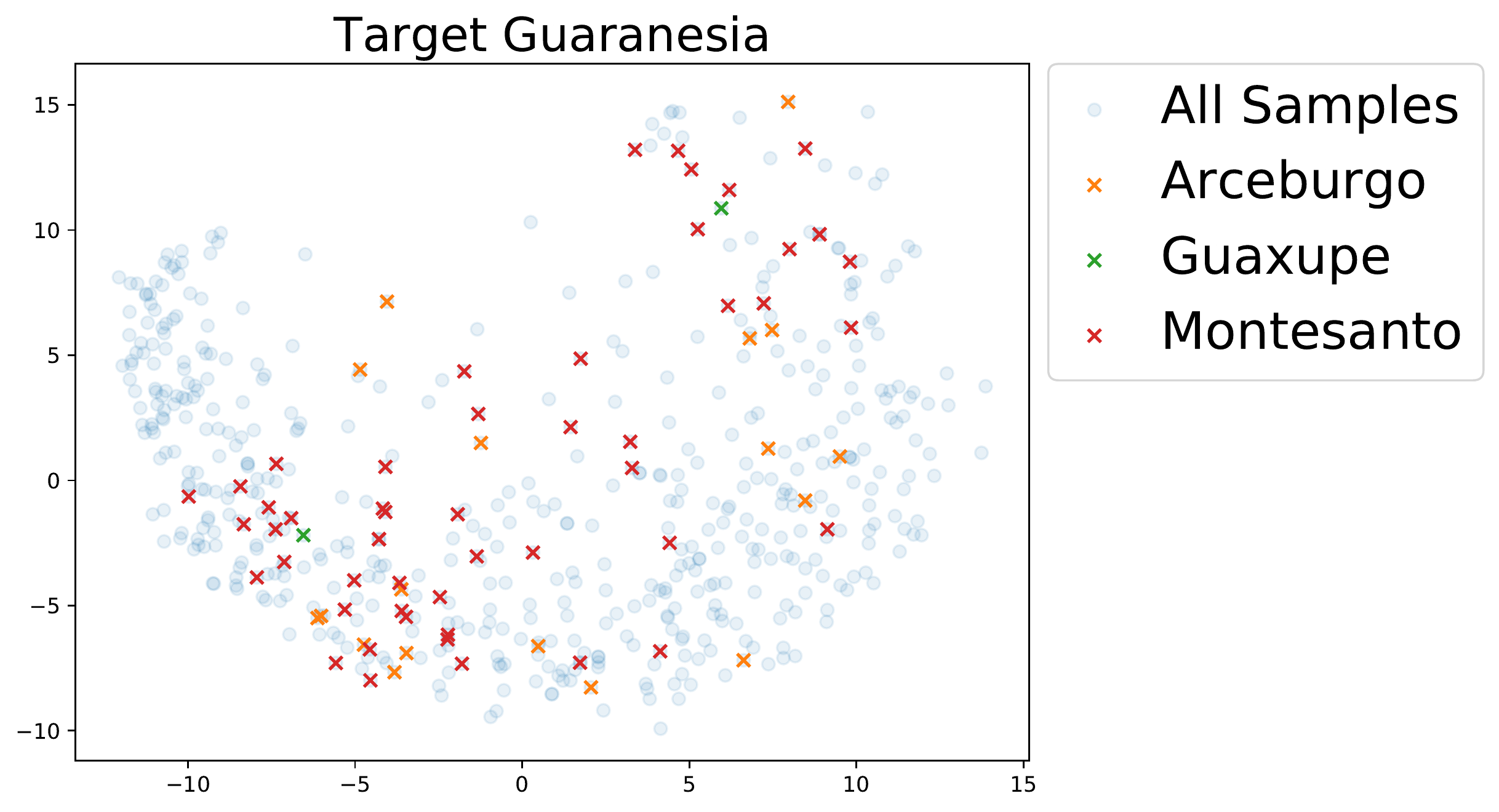}
	}
	\subfloat[Montesanto]{
		\includegraphics[width=0.48\columnwidth]{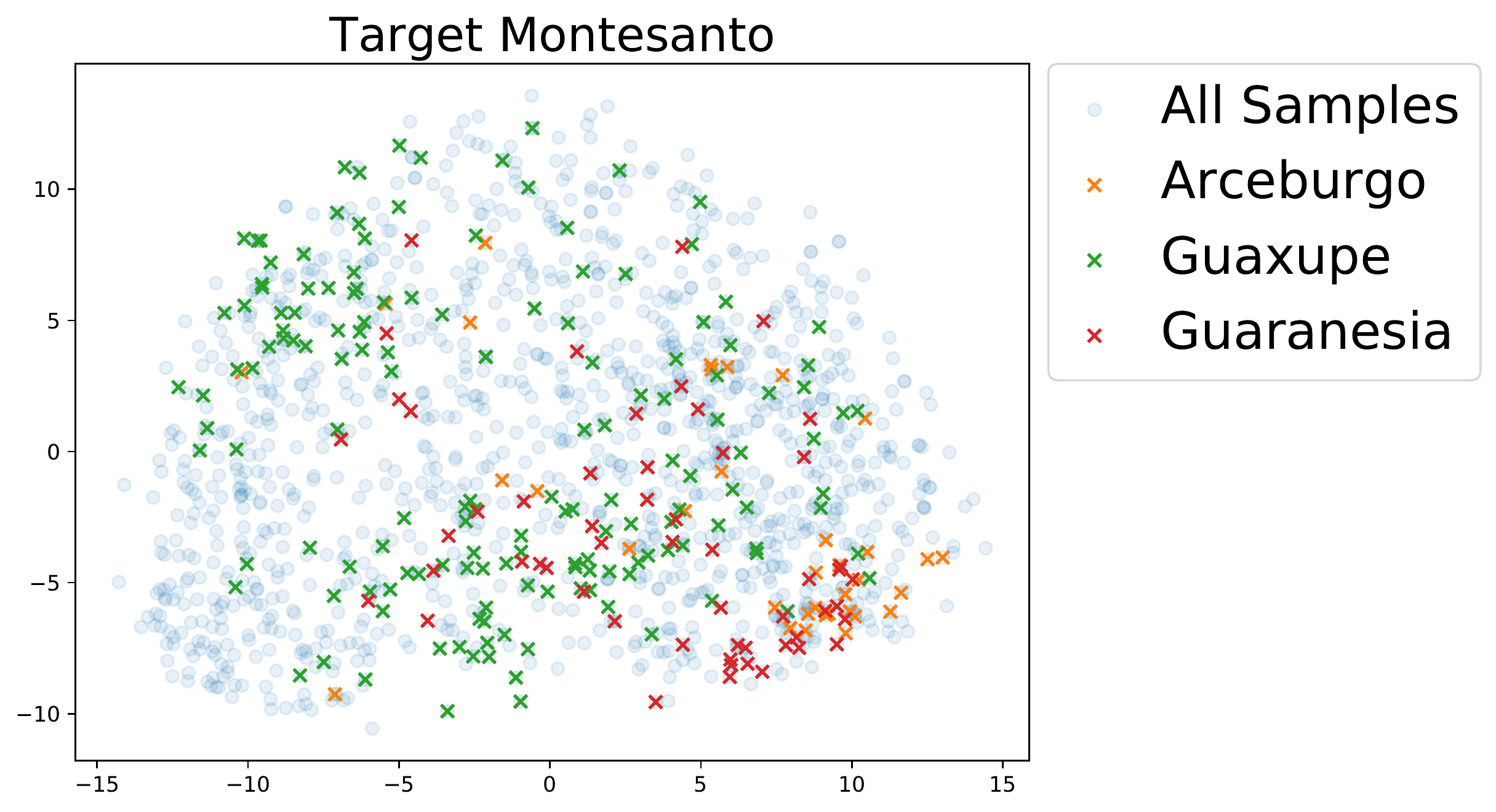}
	}
	\caption{2D-Space Projections using PCA}
	\label{fig:complement_pca}
\end{figure}

\begin{figure}[]
	\centering
	\subfloat[Arceburgo]{
		\includegraphics[width=0.48\columnwidth]{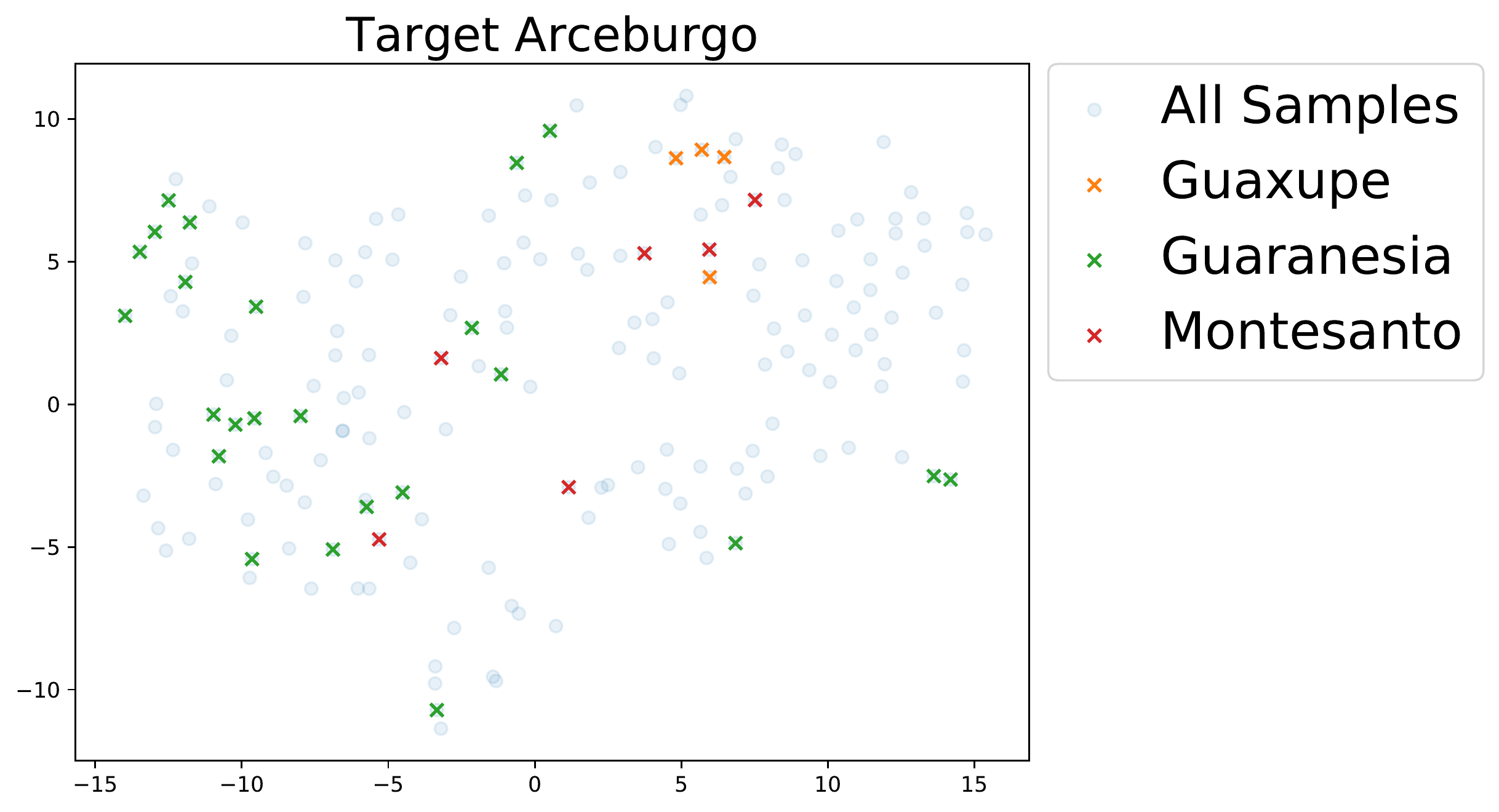}
	}
	\subfloat[Guaxup\'{e}]{
		\includegraphics[width=0.48\columnwidth]{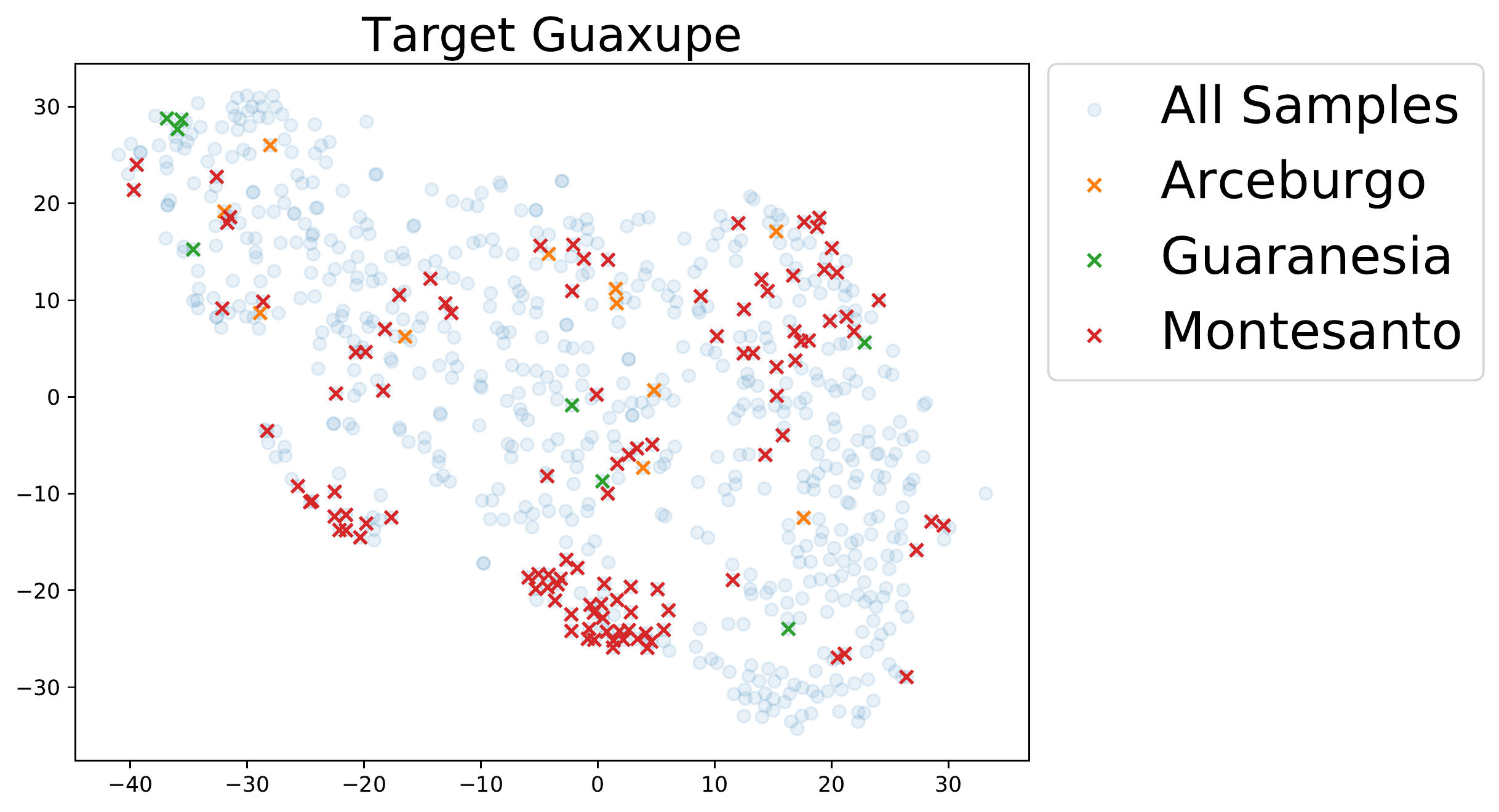}
	}
	
	\subfloat[Guaran\'{e}sia]{
		\includegraphics[width=0.48\columnwidth]{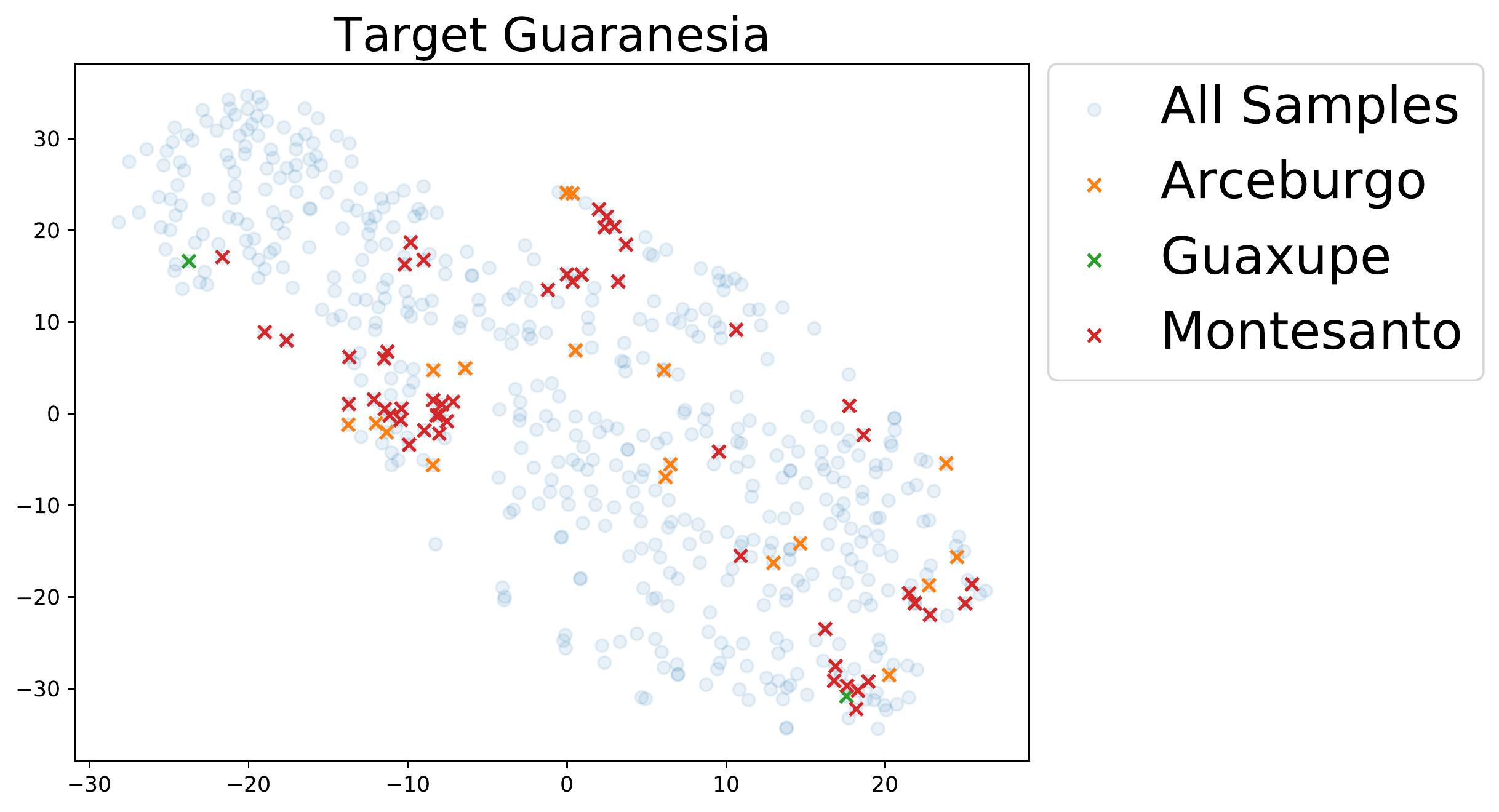}
	}
	\subfloat[Montesanto]{
		\includegraphics[width=0.48\columnwidth]{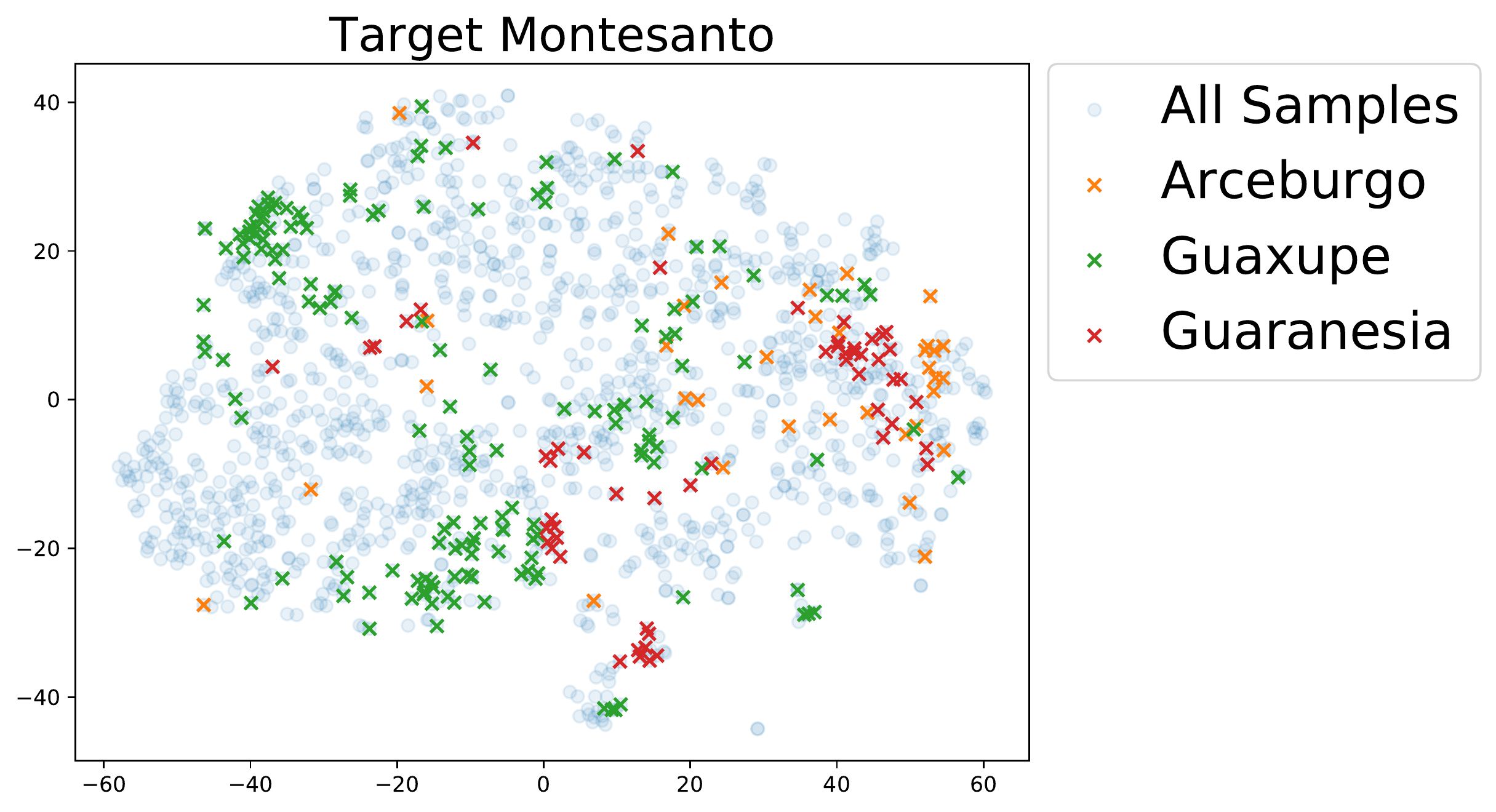}
	}
	\caption{2D-Space Projections using TSNE}
	\label{fig:complement_tsne}
\end{figure}

In this section we investigate the spatial relationship among the samples. Given a fix adaptation approach, we are focusing on samples that were correctly predicted exclusively for that source data in specific. For this purpose, we propose a visual analysis of these samples using two different methods to project the original representation of data in 2D-space: Principal Component Analysis and t-Distributed Stochastic Neighbor Embedding (TSNE). The projections from PCA and TSNE data are showed in Figure \ref{fig:complement_pca} and \ref{fig:complement_tsne} respectively.

With a visual analysis of the projections, it is possible to notice important aspects of data and the complementariness between source data. First, PCA projections show little insight into the spatial relationship of correctly predicted samples; instead, it shows sparsity over the features space. The PCA projection is a powerful dimensional reduction technique since it projects the original high-dimensional data in a low-dimensional space preserving the maximum variance as possible. However, PCA not preserve the local structure of original data, i.e., points that are close, regarding some metric, in original high-dimensional space do not remain close in the new low-dimensional space. Second, in contrast with PCA, TSNE using a non-linear manifold approach can successfully create a low-dimensional representations preserving local structures, as shown in Figure \ref{fig:complement_tsne}. In addition, we can notice a leaning of a complementariness between learning models, since the samples corrected predict from different sources are tending to create clusters. This behavior in projections can be a suggestive interpretation of shared properties between the source and target data where the clusters show samples whose are more likely to be drawn from a specific source data. Another way of seeing the previous interpretation is taking in consideration the fact of remote sensing images can present a high intra-class variance due to the huge spatial extension explored. An entire image can be seen as a composition of several probabilities distributions which some of them are better explained from different sources of data.

\section{Conclusion}


In this paper, we have conducted a comparative experimental analysis of seven UDA approaches to perform automatic coffee crop mapping. 
We conducted three sets of experiments with the intent of verifying whether existing approaches to unsupervised domain adaptation can assist in the transfer of knowledge between datasets of different geographic domains. 


The main conclusion is that employ an UDA strategy is more effective than perform transfer knowledge without any adaptation.
Experimental results also showed a great sensitive of the methods compared over different normalization pre-processing steps. In terms of mean accuracy, the Transfer Component Analysis (TCA)~\cite{pan2011domain} presents the must suitable results. In addition, the negative transfer phenomenon is noticed in several adaptation combinations supporting the importance of an effective adaptation. Analyzing the complementarity of predictions, was showed an existence of additional information that could be exploited from multiple source data to build a more reliable learning model. At last, in visual analysis was possible to identify a formation of clusters betweens samples correct predicted using different source data. This observation shows that some samples from target data are likely to be drawn from specific source. This inspection indicate that a robust UDA approach needs to recognize the importance of multiples sources, considering that each source data have a different contribution for distinct samples from the target.
	
As future work, we intend to investigate ways for avoiding negative transfer and employ UDA strategies in other vegetation mapping applications.

\bibliographystyle{IEEEtran}
\bibliography{bibliography}

\begin{thebibliography}{10}
\providecommand{\url}[1]{#1}
\csname url@samestyle\endcsname
\providecommand{\newblock}{\relax}
\providecommand{\bibinfo}[2]{#2}
\providecommand{\BIBentrySTDinterwordspacing}{\spaceskip=0pt\relax}
\providecommand{\BIBentryALTinterwordstretchfactor}{4}
\providecommand{\BIBentryALTinterwordspacing}{\spaceskip=\fontdimen2\font plus
\BIBentryALTinterwordstretchfactor\fontdimen3\font minus
  \fontdimen4\font\relax}
\providecommand{\BIBforeignlanguage}[2]{{%
\expandafter\ifx\csname l@#1\endcsname\relax
\typeout{** WARNING: IEEEtran.bst: No hyphenation pattern has been}%
\typeout{** loaded for the language `#1'. Using the pattern for}%
\typeout{** the default language instead.}%
\else
\language=\csname l@#1\endcsname
\fi
#2}}
\providecommand{\BIBdecl}{\relax}
\BIBdecl

\bibitem{SantosPenattiTorres:visapp2010}
J.~A. dos Santos, O.~A.~B. Penatti, and R.~da~S.~Torres, ``Evaluating the
  potential of texture and color descriptors for remote sensing image retrieval
  and classification,'' in \emph{VISAPP}, Angers, France, May 2010.

\bibitem{Nogueira2015:CIARP}
K.~Nogueira, W.~R. Schwartz, and J.~A. dos Santos, ``Coffee crop recognition
  using multi-scale convolutional neural networks,'' in \emph{CIARP}, 2015, pp.
  67--74.

\bibitem{nogueira2017towards}
K.~Nogueira, O.~A. Penatti, and J.~A. dos Santos, ``Towards better exploiting
  convolutional neural networks for remote sensing scene classification,''
  \emph{Pattern Recognition}, vol.~61, pp. 539--556, 2017.

\bibitem{zhangtransfer}
J.~Zhang, W.~Li, and P.~Ogunbona, ``Transfer learning for cross-dataset
  recognition: A survey,'' \emph{arXiv preprint arXiv:1705.04396}, 2017.

\bibitem{pan2011domain}
S.~J. Pan, I.~W. Tsang, J.~T. Kwok, and Q.~Yang, ``Domain adaptation via
  transfer component analysis,'' \emph{IEEE Transactions on Neural Networks},
  vol.~22, no.~2, pp. 199--210, 2011.

\bibitem{long2013transfer}
M.~Long, J.~Wang, G.~Ding, J.~Sun, and P.~S. Yu, ``Transfer feature learning
  with joint distribution adaptation,'' in \emph{ICCV}, 2013, pp. 2200--2207.

\bibitem{long2014transfer}
------, ``Transfer joint matching for unsupervised domain adaptation,'' in
  \emph{CVPR}, 2014, pp. 1410--1417.

\bibitem{fernando2013unsupervised}
B.~Fernando, A.~Habrard, M.~Sebban, and T.~Tuytelaars, ``Unsupervised visual
  domain adaptation using subspace alignment,'' in \emph{ICCV}, 2013, pp.
  2960--2967.

\bibitem{gong2012geodesic}
B.~Gong, Y.~Shi, F.~Sha, and K.~Grauman, ``Geodesic flow kernel for
  unsupervised domain adaptation,'' in \emph{CVPR}.\hskip 1em plus 0.5em minus
  0.4em\relax IEEE, 2012, pp. 2066--2073.

\bibitem{sun2016return}
B.~Sun, J.~Feng, and K.~Saenko, ``Return of frustratingly easy domain
  adaptation.'' in \emph{AAAI}, vol.~6, no.~7, 2016, p.~8.

\bibitem{zhang2017joint}
J.~Zhang, W.~Li, and P.~Ogunbona, ``Joint geometrical and statistical alignment
  for visual domain adaptation,'' in \emph{CVPR}, 2017, pp. 1859--1867.

\bibitem{stehling2002compact}
R.~O. Stehling, M.~A. Nascimento, and A.~X. Falc{\~a}o, ``A compact and
  efficient image retrieval approach based on border/interior pixel
  classification,'' in \emph{CIKM}.\hskip 1em plus 0.5em minus 0.4em\relax ACM,
  2002, pp. 102--109.

\end{thebibliography}

\end{document}